\pdfoutput=1

\documentclass[11pt]{article}
\usepackage[dvipsnames, svgnames, x11names]{xcolor}

\usepackage{authblk}
\usepackage{EMNLP2023}

\usepackage{times}
\usepackage{latexsym}

\usepackage[T1]{fontenc}

\usepackage[utf8]{inputenc}

\usepackage{microtype}

\usepackage{inconsolata}

\usepackage{booktabs}
\usepackage{makecell}
\usepackage{pifont}
\usepackage{bm}
\usepackage{amsmath}
\usepackage{multirow}
\usepackage{graphicx}

\graphicspath{{./figures/}}

\title{Grammatical Error Correction via Mixed-Grained Weighted Training}

\author{Jiahao Li$^1$, Quan Wang$^2$\thanks{\hspace{0.15cm}Corresponding author: Quan Wang.} , Chiwei Zhu$^1$, Zhendong Mao$^1$,  Yongdong Zhang$^1$\\ \\
  $^1$University of Science and Technology of China, Hefei, China \\
  $^2$MOE Key Laboratory of Trustworthy Distributed Computing and Service, \authorcr Beijing University of Posts and Telecommunications, Beijing, China \\
  {\tt jiahao66@mail.ustc.edu.cn, wangquan@bupt.edu.cn} \\
  {\tt tanz@mail.ustc.edu.cn, zdmao@ustc.edu.cn, zhyd73@ustc.edu.cn}}

\begin{document}
\maketitle
\begin{abstract}
The task of Grammatical Error Correction (GEC) aims to automatically correct grammatical errors in natural texts.
Almost all previous works treat annotated training data equally, but inherent discrepancies in data are neglected. 
In this paper, the inherent discrepancies are manifested in two aspects, namely, accuracy of data annotation and diversity of potential annotations. 
To this end, we propose MainGEC, which designs token-level and sentence-level training weights based on inherent discrepancies in accuracy and potential diversity of data annotation, respectively, and then conducts mixed-grained weighted training to improve the training effect for GEC.
Empirical evaluation shows that whether in the Seq2Seq or Seq2Edit manner, MainGEC achieves consistent and significant performance improvements on two benchmark datasets, demonstrating the effectiveness and superiority of the mixed-grained weighted training. 
Further ablation experiments verify the effectiveness of designed weights of both granularities in MainGEC.

\end{abstract}

\section{Introduction}

The task of Grammatical Error Correction (GEC) aims to automatically correct grammatical errors in natural texts, which is extremely beneficial for language learners, such as children and non-native speakers \citep{DBLP:journals/corr/abs-2211-05166}.
The currently dominant neural GEC methods are categorized into two groups, {\it i.e.}, Seq2Seq methods and Seq2Edit methods. 
Seq2Seq methods treat GEC as a monolingual translation task, regarding errorful sentences as the source language and error-free sentences as the target language \citep{DBLP:conf/naacl/YuanB16,DBLP:conf/acl/SunGWW20,DBLP:conf/emnlp/0004ZLBLZ22}.
Seq2Edit methods treat GEC as a sequence tagging task, which predicts a tagging sequence of edit operations to perform correction \citep{DBLP:conf/emnlp/AwasthiSGGP19,DBLP:conf/bea/OmelianchukACS20,DBLP:conf/acl/TarnavskyiCO22}.

\begin{table}[ht]
\small
\centering\setlength{\tabcolsep}{4pt}
\begin{tabular*}{0.48 \textwidth}{@{\extracolsep{\fill}}ll}
\toprule
\multicolumn{2}{@{}l}{\textbf{Accuracy of Data Annotation}}  \\
\toprule
Sample 1: &  \makecell[l]{$\!\!\!\!$	Their new house near the beach is very nice .}   \\ 
Annotation:  & \makecell[l]{$\!\!\!\!$   \textcolor{red}{Your} (\textcolor{red}{\ding{55}}) new house near the beach is very nice .} \\
\midrule
Sample 2: & $\!\!\!\!$	I read your email yesterday but I had n't had\\ 
& $\!\!\!\!$ the time to reply yet . \\
Annotation:  & $\!\!\!\!$   I read your email yesterday but I \textcolor{blue}{have} (\textcolor{blue}{\ding{51}}) n't \\
& $\!\!\!\!$ had the time to reply \textcolor{red}{until now} (\textcolor{red}{\ding{55}}) . \\
\midrule
Sample 3: & \makecell[l]{$\!\!\!\!$	do you have best friend in your life ? }  \\
Annotation:  & \makecell[l]{$\!\!\!\!$   \textcolor{blue}{Do} (\textcolor{blue}{\ding{51}}) you have \textcolor{blue}{a best friend} (\textcolor{blue}{\ding{51}}) in your life ?} \\
\bottomrule
\toprule
\multicolumn{2}{@{}l}{\textbf{Diversity of Potential Annotations}}  \\
\toprule
Sample 4: & $\!\!\!\!$	Natural environment destroyed that is a people \\
&$\!\!\!\!$ focus on frequently problem .  \\
Annotation:  & $\!\!\!\!$   The natural environment is being destroyed \textcolor{LimeGreen}{.} \\
& $\!\!\!\!$ \textcolor{LimeGreen}{That} is a problem people focus on frequently . \\
Alternative:  & $\!\!\!\!$   The natural environment is being destroyed \textcolor{LimeGreen}{,} \\
& $\!\!\!\!$ \textcolor{LimeGreen}{which} is a problem people focus on frequently . \\
\midrule
Sample 5: & $\!\!\!\!$	Secondly there are not much variety of dessert \\
& $\!\!\!\!$ mainly fruits and puddings .   \\
Annotation:  & $\!\!\!\!$   Secondly , there \textcolor{LimeGreen}{is not a lot of variety in the}\\
& $\!\!\!\!$ \textcolor{LimeGreen}{desserts ,} mainly fruits and puddings . \\
Alternative:  & $\!\!\!\!$   Secondly , there \textcolor{LimeGreen}{are not many varieties of}\\
& $\!\!\!\!$ \textcolor{LimeGreen}{desserts ,} mainly fruits and puddings . \\
\midrule
Sample 6: & $\!\!\!\!$	One of my favourite books are Diary of a\\
& $\!\!\!\!$ Wimpy Kid .  \\
Annotation:  & $\!\!\!\!$   One of my favourite books \textcolor{LimeGreen}{is} Diary of a\\
&$\!\!\!\!$ Wimpy Kid . \\
\bottomrule
\end{tabular*}
\caption{\label{motivation}Instances from the BEA-19 \citep{DBLP:conf/bea/BryantFAB19} training set to show the discrepancies in the annotated training data. Erroneous annotations are in red, correct annotations are in blue, and multiple potential annotations are in green.} 
\end{table}

Whether in the Seq2Seq or Seq2Edit manner, almost all previous works treat annotated training data equally \citep{DBLP:conf/acl/RotheMMKS20,DBLP:conf/acl/TarnavskyiCO22}, that is, assigning the same training weight to each training sample and each token therein. 
However, inherent discrepancies in data are completely neglected, causing degradation of the training effect. 
Specifically, inherent discrepancies may be manifested in two aspects, {\it i.e.},  accuracy of data annotation and diversity of potential annotations. 
The discrepancy in \textbf{accuracy of data annotation} refers to the uneven annotation quality, which is caused by differences in the annotation ability of annotators and the difficulty of samples \citep{DBLP:conf/naacl/0004LBLZLHZ22}. 
For example, in Table~\ref{motivation}, Sample 1 and Sample 2 contain annotation errors to varying degrees, while the annotation of Sample 3 is completely correct.
The discrepancy in \textbf{diversity of potential annotations} refers to the different amounts of potential reasonable alternatives to annotation. 
Usually, it differs due to different sentence structures or synonymous phrases. 
For example, Sample 4 and 5 potentially have multiple reasonable annotations, while Sample 6 probably only has a single reasonable annotation. 
Due to the above data discrepancies, training data should be distinguished during the training process, by being assigned well-designed weights.

In this paper, we propose \textbf{MainGEC} ({\it i.e.}, \textbf{M}ixed-gr\textbf{a}ined we\textbf{i}ghted trai\textbf{n}ing for \textbf{GEC}), which designs mixed-grained weights for training data based on inherent discrepancies therein to improve the training effect for GEC.
First, we use a well-trained GEC model (called a teacher model) to quantify accuracy and potential diversity of data annotation. 
On the one hand, the accuracy of annotations is estimated by the generation probability of the teacher model for each target token, which represents the acceptance degree of the teacher model for the current annotation.
Then, the quantified accuracy is converted into token-level training weights, as the accuracy of annotations may vary not only across samples but even across tokens in a single sample, {\it e.g.}, sample 2 in Table~\ref{motivation}.
On the other hand,  the diversity of potential annotations is estimated by the information entropy of output distribution of the teacher model for each training sample, which actually represents the uncertainty, {\it i.e.}, diversity, of the target sentences that the teacher model is likely to generate.
Then, the quantified potential diversity is converted into sentence-level training weights, considering that the potential annotations may involve the semantics and structures of the entire sentence. 
Finally, the token-level and sentence-level weigths constitute our mixed-grained weights for the training process.

\citet{DBLP:journals/tacl/LichtargeAK20} also considers to allocate training weights for samples. 
However, they only consider discrepancies in synthetic data and still treat human-annotated data equally, while the discrepancies we consider are across all data.
Additionally, they only design sentence-level weighting, without token-level weighting considered in this paper. 
From another perspective, our method can be regarded as an "alternative" knowledge distillation method. 
Compared to \citet{xia2022chinese} applying general knowledge distillation on GEC, our method uses a teacher model to obtain mixed-grained training weights based on inherent discrepancies in data to guide the training process, rather than forcing the output distribution of the student model to be consistent with that of the teacher model.

We apply our mixed-grained weighted training to the mainstream Seq2Seq and Seq2Edit methods, and both of them achieve consistent and significant performance improvements on two benchmark datasets, verifying the superiority and generality of the method.
In addition, we conduct ablation experiments, further verifying the effectiveness of the designed weights of both granularities.
Besides, we conduct the comparative experiment with the general knowledge distillation method on GEC, verifying that our mixed-grained training weighting strategy outperforms the general knowledge distillation strategy. 

The main contributions of this paper are summarized as follows: 
(1) We investigate two kinds of inherent discrepancies in data annotation of GEC for the first time, and propose MainGEC, which designs mixed-grained training weights based on the discrepancies above to improve the training effect. 
(2) The extensive empirical results show that MainGEC achieves consistent and significant performance improvements over the mainstream Seq2Seq and Seq2Edit methods on two benchmarks, proving the effectiveness and generality of our method for GEC.

\section{Preliminary}

This section presents the formulation of GEC task and currently mainstream Seq2Seq and Seq2Edit methods for GEC. 

\begin{figure*}[t]
\centering
\includegraphics[scale=0.45]{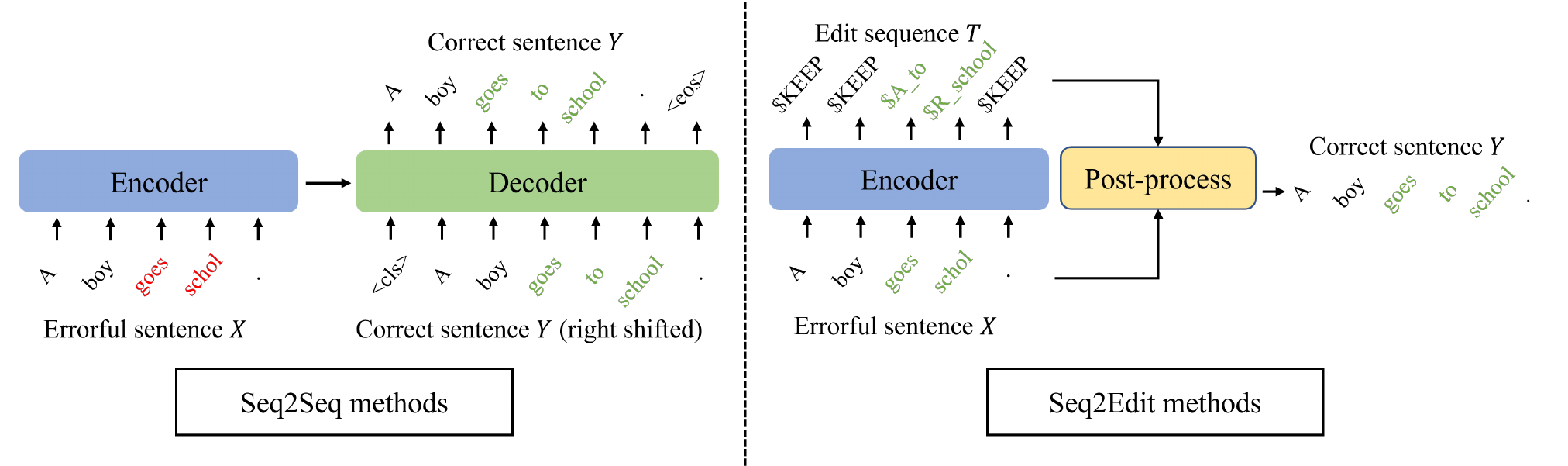}
\caption{Overview of Seq2Seq and Seq2Edit methods for GEC. \textbf{Left:} Seq2Seq methods  encode the errorful sentence $X$ by an encoder, and autoregressively generate the corresponding correct sentence $Y$ via a decoder. \textbf{Right:} Seq2Edit methods employ a sequence tagging model to predict a tagging sequence $T$ of edit operations corresponding to the errorful sentence $X$, and the correct sentence $Y$ is obtained by applying editing operations to $X$ via post-processing. Here, the tag {\it \$A\_to} denotes appending a new token "to" next to the current token "goes", and the tag {\it \$R\_school} denotes replacing the current token "schol" with "school". }
\label{fig:1}
\end{figure*}

\subsection{Problem Formulation}
Grammatical Error Correction (GEC) is to correct grammatical errors in natural texts.
Given an errorful sentence $X = \{x_1, x_2, \cdots, x_m\}$ with $m$ tokens, a GEC system takes $X$ as input, corrects grammatical errors therein, and outputs a corresponding error-free sentence $Y = \{y_1, y_2,$ $\cdots,  y_n \} $ with $n$ tokens.  
In general, the target sentence $Y$ often has substantial overlap with the source sentence $X$. 

\subsection{Seq2Seq Methods}

The Seq2Seq methods employ the encoder-decoder framework, where the encoder encodes the entire errorful sentence $X$ into corresponding hidden states, and the decoder autoregressively generates each token in $Y$ based on the hidden states and the previously generated tokens, as shown on the left in Figure \ref{fig:1}. 

The general objective function of the Seq2Seq methods is to minimize the negative log-likelihood loss:
\begin{equation*}\label{eq:seq2seq}
\mathcal{L}(\theta) = - \sum_{i=1}^{n} \log p(\hat{y}_i=y_i|X, Y_{<i}, \theta),
\end{equation*}
where $\theta$ is learnable model parameters, $\hat{y}_i$ is the $i$-th token predicted by the model, and $Y_{<i} = \{y_1, y_2,$ $\cdots,  y_{i-1} \} $ denotes a set of tokens before the  $i$-th token $y_i$.

\subsection{Seq2Edit Methods}

Due to the substantial overlap between $X$ and $Y$, autoregressive generation for the entire target $Y$ is inefficient, and Seq2Edit methods is a good alternative. 
The Seq2Edit methods usually employ a sequence tagging model made up of a BERT-like encoder stacked with a simple classifier on the top, as shown on the right in Figure \ref{fig:1}. 
At first, a pre-defined set of tags is required to denote edit operations. 
In general, this set of tags contains universal edits, ({\it e.g.} {\it \$KEEP} for keeping the current token unchanged, {\it \$DELETE} for deleting the current token, {\it \$VERB\_FORM} for conversion of verb forms, {\it etc})\footnote{Here we take GECToR's tags \citep{DBLP:conf/bea/OmelianchukACS20} for illustration. } and token-dependent edits, ({\it e.g.} {\it \$APPEND\_$e_i$} for appending a new token $e_i$ next to the current token, {\it \$REPLACE\_$e_i$} for replacing the current token with another token $e_i$). 
Considering the linear growth of tag vocab's size taken by token-dependent edits, usually, a moderate tag vocab's size is set to balance edit coverage and model size based on the frequency of edits.
Then, the original sentence pair $(X, Y)$ is converted into a sentence-tags pair $(X, T)$ of equal length.
Specifically, the target sentence $Y$ is aligned to the source sentence $X$ by minimizing the modified Levenshtein distance, and then converted to a tag sequence $T = \{t_1, t_2,$ $\cdots,  t_m \} $. Refer to \citet{DBLP:conf/bea/OmelianchukACS20} for more details. 

In training, the general objective function of the Seq2Edit methods is to minimize the negative log-likelihood loss for the tag sequence:
\begin{equation*}\label{eq:seq2edit}
\mathcal{L}^{s2e}(\theta) = - \sum_{i=1}^{m} \log p(\hat{t}_i=t_i|X, \theta),
\end{equation*}
where $\hat{t}_i$ is the $i$-th tag predicted by the model.
During inference, Seq2Edit methods predict a tagging sequence $\hat{T}$ at first, and then apply the edit operations in the source sentence $X$ via post-processing to obtain the predicted result $\hat{Y}$.

\section{Our Approach}
\begin{figure}[t]
\centering
\includegraphics[scale=0.5]{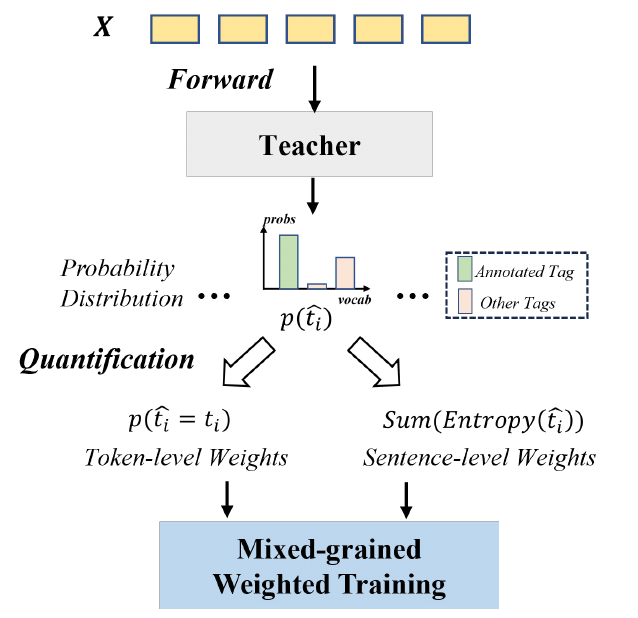}
\caption{Illustration of MainGEC. MainGEC converts the target distribution generated by a teacher model and original targets into mixed-grained weights, and conducts weighted training with them. }
\label{fig:2}
\end{figure}

This section presents our approach, MainGEC which designs mixed-grained weights for training data based on inherent discrepancies therein to improve the training effect for GEC. 
Below, we first elaborate on how to quantify accuracy and potential diversity of data annotation at the token-level and sentence-level respectively, and convert quantified features to training weights of both granularities, correspondingly. 
Then, based on both-grained weights, the overall mixed-grained weighted training strategy is introduced.
Figure \ref{fig:2} summarizes the overall architecture of MainGEC.

\subsection{Token-Level Weights}

Due to differences in the annotation ability of annotators and the difficulty of samples, there is a discrepancy in the accuracy of data annotation. 
Actually, this discrepancy exists not only across samples but even across tokens in a single sample. 
To this end, a well-trained GEC model is used to quantify the accuracy of data annotation for each token in all training samples, and then they are converted into token-level training weights.

\paragraph{For Seq2Seq Methods}
The source sentence $X$ is fed into a well-trained Seq2Seq GEC model (called the teacher model), and the accuracy of the data annotation is estimated by the generation probability of the teacher model for each target token $y_i$ :
\begin{equation*}\label{eq:seq2seq accuracy}
\bm{Acc}(y_i) = p(\hat{y}_i=y_i|X, Y_{<i}, \theta_{T}), 
\end{equation*}
where $i \in \{1, 2, \cdots, n \}$, $\theta_{T}$ is parameters of the teacher model. 
Actually, this estimation implies the extend to which the teacher model agrees with the current annotation, which can be a measure of the accuracy.
Then, quantified accuracy of data annotation for each target token can be directly regarded as the token-level training weight, as the higher accuracy of data annotation means the better annotation quality and thus a higher token-level training weight should be assigned for training.
The token-level training weights for Seq2Seq methods is defined as:
\begin{equation*}\label{eq:seq2seq token}
\bm{w}_{token}(y_i) = \bm{Acc}(y_i).
\end{equation*}

\paragraph{For Seq2Edit Methods}
Similarly, the accuracy of the data annotation is estimated by the generation probability of a well-trained Seq2Edit teacher model for each target tag $t_i$:
\begin{equation*}\label{eq:seq2edit accuracy}
\bm{Acc}(t_i) = p(\hat{t}_i=t_i|X, \theta_{T}),
\end{equation*}
where $i \in \{1, 2, \cdots, m \}$. 
Correspondingly, the token-level training weights for each target tag is defined as:
\begin{equation*}\label{eq:seq2edit token}
\bm{w}_{token}(t_i) = \bm{Acc}(t_i).
\end{equation*}

\subsection{Sentence-Level Weigths}

Due to different sentence structures or synonymous phrases, there can be multiple potential reasonable alternatives to the single target sentence $Y$ of a training sample $(X, Y)$. 
Further, the amounts of potential reasonable alternatives may differ across all samples, which is referred to as the discrepancy in the diversity of potential annotations.
Therefore, we quantify the diversity of potential annotations for each training sample by the same teacher model above, and convert them into sentence-level training weights.

\paragraph{For Seq2Seq Methods}
We feed the source sentence $X$ into the teacher model to obtain the probability distribution of its prediction result. 
For this sample $(X, Y)$, the diversity of potential annotations is estimated by the information entropy of this distribution:
\begin{equation*}\label{eq:seq2seq diversity}
\bm{Div}(X, Y) = \frac{1}{n} \sum_{i=1}^{n} \frac{\bm{H}(\hat{y}_i|X, Y_{<i}, \theta_{T})}{log|V|},
\end{equation*}
where $|V|$ is the vocab size and $\bm{H}()$ denotes the entropy of a random variable, with $log|V|$ for normalization. 
Here, lower information entropy means that the teacher model produces a sparser and sharper probability distribution.
This leads to the fact that fewer candidate target sentences are likely to be generated, {\it i.e.}, there is less diversity of potential annotations therein.
Further, this means the teacher model has more confidence for the target annotation, and a higher sentence-level training weight should be assigned during training.
Therefore, a monotonically decreasing function and proper boundary processing are applied to the quantified diversity of potential annotations to obtain the sentence-level training weight for the sample $(X, Y)$:
\begin{equation*}\label{eq:seq2seq sentence}
\bm{w}_{sent}(X, Y) = \rm{Max}[\frac{\log (\bm{Div}(X, Y) + \epsilon)}{\log \epsilon}, \epsilon],
\end{equation*}
where $\epsilon$ is a small positive quantity ({\it e.g.}, $e^{-9}$).

\paragraph{For Seq2Edit Methods}
Similarly, the diversity of potential annotations is estimated by the information entropy of output distribution of a Seq2Edit teacher model for a sample $(X, T)$:
\begin{equation*}\label{eq:seq2edit diversity}
\bm{Div}(X, T) = \frac{1}{m} \sum_{i=1}^{m} \frac{\bm{H}(\hat{t}_i|X, \theta_{T})}{log|E|},
\end{equation*}
where $|E|$ is the size of the pre-defined tag set. 
Correspondingly, the sentence-level training weight for the sample $(X, T)$ is defined as:
\begin{equation*}\label{eq:seq2edit sentence}
\bm{w}_{sent}(X, T) = \rm{Max}[\frac{\log (\bm{Div}(X, T) + \epsilon)}{\log \epsilon}, \epsilon].
\end{equation*}

\subsection{Mixed-Grained Weighted Training}

The mixed-grained weighted training is to simply integrate both-grained weights into the training process. 
During training, the sentence-level weights determine the contribution of each sample to update the model parameters, while further token-level weights are used to adjust the importance of each token/tag therein.
\paragraph{For Seq2Seq Methods} 
We use the sentence-level and token-level weights as factors of the training loss for the samples and the tokens in them, respectively.
The overall loss function of our mixed-grained weighted training is defined as:
\begin{equation*}\label{eq:seq2seq mix}
\begin{split}
\mathcal{L}_{w}&(\theta)  = - \sum_{(X, Y) \in D} \bm{w}_{sent}(X, Y) * \\
& \sum_{i=1}^{n} \bm{w}_{token}(y_i) * \log p(\hat{y}_i=y_i|X, Y_{<i}, \theta),
\end{split}
\end{equation*}
where $D$ is all training corpus. 
\paragraph{For Seq2Edit Methods} 
Similarly, the loss function of our MainGEC for Seq2Edit methods is defined as:
\begin{equation*}\label{eq:seq2edit mix}
\begin{split}
\mathcal{L}_{w}&(\theta) = - \sum_{(X, T) \in D^T} \bm{w}_{sent}(X, T) * \\
& \sum_{i=1}^{m} \bm{w}_{token}(t_i) * \log p(\hat{t}_i=t_i|X, \theta),
\end{split}
\end{equation*}
where $D^T$ is all training data after the tag transformation.

\section{Experiments and Results}

This section introduces our experiments and results on two benchmarks, {\it i.e.}, CONLL-14 \citep{DBLP:conf/conll/NgWBHSB14} and BEA-19 \citep{DBLP:conf/bea/BryantFAB19}. 
Then, we conduct ablation experiments on both-grained training weights and comparative experiments with the general knowledge distillation method. 
Finally, a case study is presented to visualize the weights in MainGEC. 

\subsection{Experimental Setups}
\paragraph{Datasets and Evaluation Metrics}
As in \citet{DBLP:conf/acl/TarnavskyiCO22}, the training datasets we used consist of Troy-1BW \citep{DBLP:conf/acl/TarnavskyiCO22}, CLang-8\footnote{Here, CLang-8 is a clean version of Lang-8 used in \citet{DBLP:conf/acl/TarnavskyiCO22}. } \citep{DBLP:conf/acl/RotheMMKS20}, NUCLE \citep{DBLP:conf/bea/DahlmeierNW13}, FCE \citep{DBLP:conf/acl/YannakoudakisBM11}, W\&I+LOCNESS \citep{DBLP:conf/bea/BryantFAB19}. 
The statistics of the used datasets are shown in Table \ref{data}.

\begin{table}[t]
\small
\centering
\begin{tabular*}{0.48 \textwidth}{@{\extracolsep{\fill}}lrrr}
\toprule
Training Set & \#Sent & \#Tokens & \#Errors \\
\midrule
Troy-1BW  & 1.2M & 30.9M & 100\% \\
CLang-8  & 2.4M & 28.0M & 58\% \\
NUCLE  & 57K    & 1.16M & 62\%    \\
FCE  & 28K & 455K & 62\% \\
W\&I+LOCNESS  & 34K & 628K & 67\% \\
\bottomrule
\toprule
Test Set & \#Sent & \#Tokens & \#Errors \\
\midrule
CONLL-14 & 1.3K & 30.1K & 72\%    \\
BEA-19  & 4.5K & 85.7K & - \\
\bottomrule
\end{tabular*}
\caption{\label{data}Statistics of the datasets, including the number of sentences, tokens, and the proportion of errorful sentences.}
\end{table}

\begin{table*}[t]
\small
\centering\setlength{\tabcolsep}{5pt}
\begin{tabular*}{\textwidth}{@{\extracolsep{\fill}}lccccccccc}
\toprule
\multirow{2}{*}{Method} & \multirow{2}{*}{Model} & \multirow{2}{*}{Data Size}  & \multirow{2}{*}{Architecture}           & \multicolumn{3}{c}{CONLL-14}          & \multicolumn{3}{c}{BEA-19}          \\ 
\cmidrule(r){5-7}\cmidrule(l){8-10}
                            &      &          &                      & P           & R           & $F_{0.5}$           & P           & R           & $F_{0.5}$          \\ \midrule
\multirow{8}{*}{Seq2Seq}   
                            & \citet{DBLP:journals/tacl/LichtargeAK20}   & 340M  & Transformer-big & 69.4          & 43.9          & 62.1          & 67.6          & 62.5          & 66.5         \\
                            & \citet{DBLP:conf/bea/StahlbergK21}        & 540M & Transformer-big          & 72.8          & 49.5          & 66.6          & 72.1         & 64.4          & 70.4          \\
                            & T5GEC \citep{DBLP:conf/acl/RotheMMKS20}     & 2.4M & T5-large       & -          & -          & 66.1          & -          & -          & 72.1          \\
                            & T5GEC \citep{DBLP:conf/acl/RotheMMKS20}$^\ddag$     & 2.4M & T5-xxl       & -          & -          & 68.8         & -          & -          & 75.9          \\
                            & SAD \citep{DBLP:conf/acl/SunGWW20}       & 300M  & BART (12+2)        & -          & -          & 66.4          & -          & -          & 72.9          \\
                            & BART \citep{DBLP:conf/emnlp/0004ZLBLZ22} & 2.4M  & BART & 73.6          & 48.6          & 66.7          & 74          & 64.9          & 72.0        \\
                            & SynGEC \citep{DBLP:conf/emnlp/0004ZLBLZ22} & 2.4M  & BART + DepGCN & 74.7         & 49.0          & 67.6          & 75.1          & 65.5          & 72.9         \\
                            
                            \cmidrule(l){2-10}
                            & BART (reimp)$^\dag$           & \multirow{2}{*}{2.4M}  & \multirow{2}{*}{BART}                  & 74.3         & \textbf{47.7} & 66.8             & 78.1 & 58.9 & 73.3 \\ 
                            & MainGEC (BART)$^\dag$        &   &                      & \textbf{77.3} & 45.4              & \textbf{67.8} & \textbf{78.9} & \textbf{59.5} & \textbf{74.1} \\ 
\bottomrule
\toprule
\multirow{7}{*}{Seq2Edit} & PIE \citep{DBLP:conf/emnlp/AwasthiSGGP19}      & 1.2M  & BERT-large       & 66.1         & 43.0          & 59.7          & -          & -          & -          \\
                        & GECToR \citep{DBLP:conf/bea/OmelianchukACS20}      & 10.2M  & XLNET-base        & 77.5          & 40.1          & 65.3         & 79.2          & 53.9          & 72.4          \\
                        & TMTC \citep{DBLP:conf/acl/LaiZZLLCS22}  & 10.2M  & XLNET-base  & 77.9         & 41.8 & 66.4        & 81.3          & 51.6 & 72.9          \\
                        & GECToR-L \citep{DBLP:conf/acl/TarnavskyiCO22}  & 3.6M  & RoBERTa-large   & 74.4        & 41.1          & 64.0          & 80.7          & 53.4          & 73.2        \\
                        & \citet{DBLP:journals/tacl/LichtargeAK20} (reimp)   & 3.6M  & RoBERTa-large   & 76.4        & 40.5          & 64.9          & 80.4          & 54.4          & 73.4        \\
                        \cmidrule(l){2-10}
                        & GECToR-L (reimp)            & \multirow{2}{*}{3.6M}  & \multirow{2}{*}{RoBERTa-large}                 & 75.9		& \textbf{40.2} & 64.4   & 80.9        & 53.3         & 73.3 \\ 
                        & MainGEC (GECToR-L)           &   &                   & \textbf{78.9} & 39.4          & \textbf{65.7}          & \textbf{82.7} & \textbf{53.8}          & \textbf{74.5} \\ 
\bottomrule   
\end{tabular*}
\caption{\label{main result}Performance on the test sets of CONLL-14 and BEA-19, where precision (P), recall (R), $F_{0.5}$ ($F_{0.5}$) are reported ($\%$). Baseline results are directly taken from their respective literatures. Results marked by ``$\dag$'' are obtained by applying a decoding approach \citep{DBLP:conf/acl/0013W22} to adjust the precision-recall trade-off of inference, while the result marked by ``$\ddag$'' is not comparable here because it uses a much larger model capacity (11B parameters). \textbf{Note:} Better scores in MainGEC and the directly comparable baseline are bolded. }
\end{table*}

For evaluation, we consider two benchmarks, {\it i.e.}, CONLL-14 and BEA-19. 
CONLL-14 test set is evaluated by official $M^2$ scorer \citep{DBLP:conf/conll/NgWBHSB14}, while BEA-19 dev and test sets are evaluated by ERRANT \citep{DBLP:conf/acl/BryantFB17}. 
Both evaluation metrics are precision, recall and $F_{0.5}$. 

\paragraph{Baseline Methods}

We compare MainGEC against the following baseline methods. All these methods represent current state-of-the-art on GEC, in a Seq2Seq or Seq2Edit manner. 
\paragraph{Seq2Seq Methods}
\begin{itemize}
  \item {\it \citet{DBLP:journals/tacl/LichtargeAK20}} introduces a sentence-level training weighting strategy by scoring each sample based on delta-log perplexity, $\Delta ppl$, which represents the model’s log perplexity difference between checkpoints for a single sample.
  \item {\it \citet{DBLP:conf/bea/StahlbergK21}} generates more training samples based on an error type tag in a back-translation manner for GEC pre-training. 
  \item {\it T5GEC} \citep{DBLP:conf/acl/RotheMMKS20} pretrains large multi-lingual language models on GEC, and trains a Seq2Seq model on distillation data generated by the former more efficiently.
  \item {\it SAD} \citep{DBLP:conf/acl/SunGWW20} employs an asymmetric Seq2Seq structure with a shallow decoder to accelerate training and inference efficiency of GEC.
  \item {\it BART} \citep{DBLP:conf/emnlp/0004ZLBLZ22} applies a multi-stage fine-tuning strategy on pre-trained language model BART. 
  \item {\it SynGEC} \citep{DBLP:conf/emnlp/0004ZLBLZ22} extracts dependency syntactic information and incorporates it with output features of the origin encoder.
\end{itemize}

\paragraph{Seq2Edit Methods}
\begin{itemize}
    \item {\it PIE} \citep{DBLP:conf/emnlp/AwasthiSGGP19} generates a tag sequence of edit operations and applys parallel decoding to accelerate inference.
    \item {\it GECToR} \citep{DBLP:conf/bea/OmelianchukACS20} defines a set of token-level transformations and conducts 3-stage training on a tagging model.
    \item {\it TMTC} \citep{DBLP:conf/acl/LaiZZLLCS22} customizes the order of the training data based on error type, under GECToR's framework.
    \item {\it GECToR-L} \citep{DBLP:conf/acl/TarnavskyiCO22} applys Transfomer-based encoders of large configurations on GECToR.
\end{itemize}

\paragraph{Implementation Details}

For the Seq2Seq implementation, BART-large \citep{DBLP:conf/acl/LewisLGGMLSZ20} is choosed as the model backbone. 
At first, we fine-tune BART with vanilla training as the teacher model with fairseq\footnote{https://github.com/pytorch/fairseq} implementation. 
For a fair comparison with SynGEC \citep{DBLP:conf/emnlp/0004ZLBLZ22}, the training scheme here is to just fine-tune BART on the collection of all training sets excluding Troy-1BW dataset, for just one stage. 
More training details are discussed in Appendix \ref{sec:appendix a}. 

For the Seq2Edit implementation, we choose GECToR-L based on RoBERTa \citep{DBLP:journals/corr/abs-1907-11692} as the model backbone. 
The checkpoint released by GECToR-L is used for the teacher model\footnote{Please refer to Appendix \ref{teacher} for effect of different teachers on MainGEC. } to generate training weights of both granularities. 
We also conduct 3-stage training as in GECToR-L.
In Stage I, the model is pretrained on the Troy-1BW dataset. 
Then, in Stage II, the model is fine-tuned on the collection of the CLang-8, NUCLE, FCE, and W\&I+LOCNESS datasets, filtered out edit-free sentences. 
In Stage III, the model is fine-tuned on the W\&I+LOCNESS dataset. 
All training hyperparameters used in MainGEC are set to their default values as in GECToR-L.
Besides, we re-implement the most closely-related work, \citet{DBLP:journals/tacl/LichtargeAK20}, based on GECToR-L for a more equitable comparison. 

All checkpoints are selected by the loss on BEA-19 (dev) and all experiments are conducted on 1 Tesla A800 with 80G memory.

\subsection{Main Results}

Table \ref{main result} presents the main results of Seq2Seq and Seq2Edit methods. 
We can see that whether in the Seq2Seq or Seq2Edit manner, MainGEC brings consistent performance improvements on both benchmarks, verifying the effectiveness of our method.
Concretely, compared to vanilla training, our mixed-grained weighted training leads to 1.0/0.8 improvements in the Seq2Seq manner, and 1.3/1.2 improvements in the Seq2Edit manner. 
In addition, MainGEC outperforms all baselines on BEA-19 benchmark, with 1.2/1.3 improvements over previous SOTAs, while it also has a comparable performance on CONLL-14 benchmark. 
These results prove the superiority of our method. 

\subsection{Ablation Study}

\begin{table}[t]
\small
\centering\setlength{\tabcolsep}{4pt}
\begin{tabular*}{0.48 \textwidth}{@{\extracolsep{\fill}}llcccccc}
\toprule
\multirow{2}{*}{Model} & \multicolumn{3}{c}{CONLL-14} & \multicolumn{3}{c}{BEA-19} \\  
\cmidrule{2-4}\cmidrule{5-7}
&  P & R & $F_{0.5}$   & P & R & $F_{0.5}$ \\
\midrule
BART   &  74.3 & 47.7 & 66.8   & 78.1 & 58.9 & 73.3\\
\cmidrule(lr){1-7}
MainGEC &  77.3 & 45.4 & 67.8   & 78.9 & 59.5 & 74.1 \\					
w/o Token &  74.3 & 48.0 & 67.0   & 79.0 & 57.6 & 73.6 \\			
w/o Sent & 74.4 & 49.6 & 67.6   & 78.1 & 61.1 & 74.0\\
\bottomrule
\toprule
GECToR-L  &  75.9 & 40.2 & 64.4   & 80.9        & 53.3         & 73.3 \\
\cmidrule(lr){1-7}
MainGEC & 78.9 & 39.4 & 65.7   & 82.7 & 53.8 & 74.5 \\					
w/o Token  &  74.4 & 43.1 & 64.9   & 81.2 & 53.1 & 73.4 \\			
w/o Sent & 74.3 & 43.8 & 65.2   & 80 & 57.2 & 74.1\\
\bottomrule
\end{tabular*}
\caption{\label{tab:ablation}Ablation results on MainGEC, with the Seq2Seq group at the top and the Seq2Edit group at the bottom. The following changes are applied to MainGEC: removing the token-level training weights (w/o Token), and removing the sentence-level training weights (w/o Sent).}
\end{table}

We also conduct ablation study on MainGEC to investigate the effects of both-grained training weights, in the Seq2Seq and Seq2Edit manners. 
Table \ref{tab:ablation} presents the ablation results. 
It is obviously observed that whether token-level or sentence-level training weights included in MainGEC, can bring a certain degree of improvement over the baseline. 
Moreover, the mixed-grained weighted training can provide more improvements on the basis of a single grained weighted training. 

\subsection{Exploration w.r.t Knowledge Distillation}

\begin{table}[t]
\small
\centering\setlength{\tabcolsep}{4pt}
\begin{tabular*}{0.48 \textwidth}{@{\extracolsep{\fill}}lcccccc}
\toprule
\multirow{2}{*}{Method} & \multicolumn{3}{c}{CONLL-14} & \multicolumn{3}{c}{BEA-19} \\  
\cmidrule(lr){2-4}\cmidrule(lr){5-7}
& P & R & $F_{0.5}$ & P & R & $F_{0.5}$ \\
\midrule
GECToR-L&  75.9 & 40.2 & 64.4   & 80.9        & 53.3         & 73.3\\
\midrule
KD & 76.9 & \textbf{40.7} & 65.3 & 81.0 & \textbf{54.4} & 73.8\\			
MainGEC  & \textbf{78.9} & 39.4 & \textbf{65.7}   & \textbf{82.7} & 53.8 & \textbf{74.5}  \\						
\bottomrule
\end{tabular*}
\caption{\label{tab:kd} Comparison between MainGEC and the general knowledge distillation method for GEC.}
\end{table}

As there is a "teacher" model used to obtain training weights in MainGEC, it is necessary to compare MainGEC with the general knowledge distillation method \citep{xia2022chinese} for GEC, refered as KD. 
In KD,  the probability distribution generated by the teacher model is regarded as a soft objective, which supervises the entire training process with the original groundtruth together. 
Here, we reimplement KD in the Seq2Edit manner, where the teacher model is the same as before and GECToR-L (RoBERTa-large) is choosed as the student model. 
The experimental result is presented in Table \ref{tab:kd}. 
As we can see, KD brings a significant improvement over the baseline, due to extra knowledge from the teacher model. 
More importantly, with the same teacher model, MainGEC outperforms KD with a considerable margin. 
This proves our our mixed-grained weighted training is superior to KD, forcing the output distribution of the student model to be consistent with that of the teacher model.




\subsection{Case Study}

\begin{figure*}[t]
\centering
\includegraphics[scale=0.5]{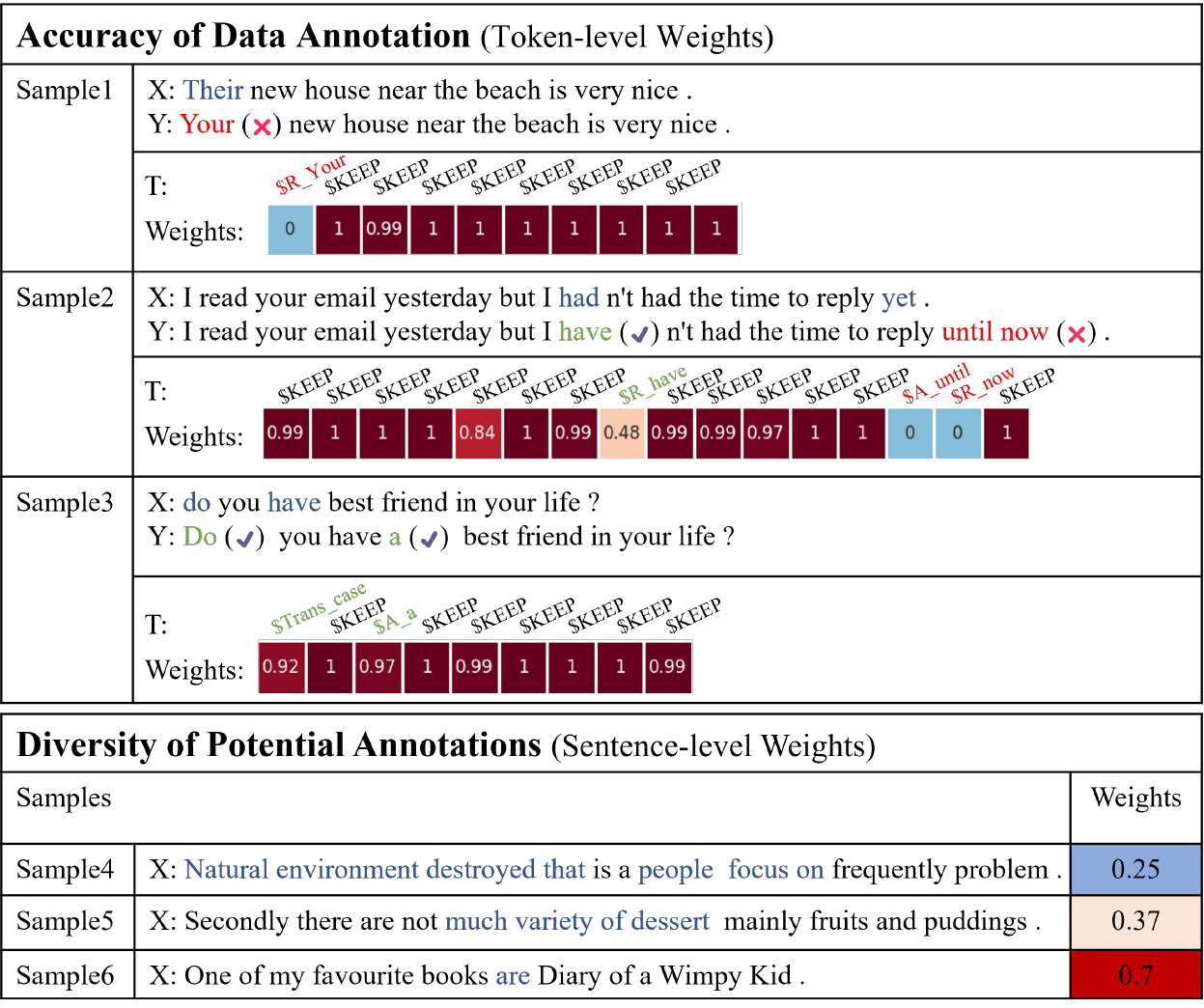}
\caption{The samples in Table \ref{motivation} and corresponding token-level or sentence-level weights obtained in MainGEC. For those token with problematic annotations or those samples with multiple potential appropriate annotations, MainGEC will assign relatively low token-level or sentence-level training weights, respectively. The correct annotations are in green, the erroneous annotations are in red, and the corresponding spans in the source sentences are in blue. }
\label{case}
\end{figure*}

Figure \ref{case} shows the same cases as in Table \ref{motivation} and their token-level or sentence-level weights obtained in MainGEC. The weights here are obtained in the Seq2Edit manner. As we can see, token-level and sentence-level weights in MainGEC indeed reflect the accuracy and potential diversity of data annotation respectively, to some extend. Specifically, for those problematic annotation, MainGEC will assign a relatively low token-level weight, and vice versa. When there are multiple potential appropriate annotations for a single sample, only one objective contained in the training set will be assigned a relatively low sentence-level weight. For example, the sentence-level weights of Sample 4 and Sample 5 in Table \ref{motivation} are relatively low due to multiple candidate sentence structures and synonymous phrases, respectively. This demonstrates that MainGEC is consistent with our motivation at first. 

\section{Related Works}
GEC is a fundamental NLP task that has received wide attention over the past decades.
Besides of the early statistical methods, the currently mainstream neural GEC methods are categorized into two groups, {\it i.e.}, Seq2Seq methods and Seq2Edit methods, in general. 

Seq2Seq methods treat GEC as a monolingual translation task, regarding errorful sentences as the source language and error-free sentences as the target language \citep{DBLP:conf/naacl/YuanB16}.
Some works \citep{DBLP:journals/corr/abs-1807-01270,DBLP:conf/ijcai/0013GM0WW22} generate considerable synthetic data based on the symmetry of the Seq2Seq's structure for data augmentation.
In addition, some works \citep{DBLP:conf/acl/KanekoMKSI20,DBLP:conf/emnlp/0004ZLBLZ22} feed additional features into the neural network to improve GEC, such as the BERT \citep{DBLP:conf/naacl/DevlinCLT19} presentation or syntactic structure of the input sentence. 

Seq2Edit methods treat GEC as a sequence tagging task, which predicts a tagging sequence of edit operations to perform correction \citep{DBLP:conf/emnlp/MalmiKRMS19}.
Parallel Iterative Edit (PIE) \citep{DBLP:conf/emnlp/AwasthiSGGP19} and GECToR \citep{DBLP:conf/bea/OmelianchukACS20} define a set of tags representing the edit operations to be modelled by their system.
\citet{DBLP:conf/acl/LaiZZLLCS22} investigates the characteristics of different types of errors in multi-turn correction based on GECToR. 
\citet{DBLP:conf/acl/TarnavskyiCO22} applies multiple ensembling methods and knowledge distillation on the large version of the GECToR system. 


\section{Conclusion}

This paper proposes MainGEC, which assigns mixed-grained weights to training data based on inherent discrepancies in data to improve the training effect for GEC.
Our method uses a well-trained GEC model to quantify the accuracy and potential diversity of data annotation, and convert them into the mixed-grained weights for the training process.
Whether in the Seq2Seq or Seq2Edit manner, MainGEC achieves consistent and significant performance improvements on two benchmark datasets, verifying the superiority and generality of the method.
In addition, further ablation experiments and comparative experiments with the general knowledge distillation method provide more insights on both-grained training weights and the perspective of knowledge distillation.

\section*{Limitations}

Our approach requires a well-trained model (called a teacher model) to obtain weights of two granularities before training. Therefore, compared to vanilla training, MainGEC has the additional preparation step to first acquire a teacher model (publicly released or trained by yourself) and then compute the weights by a forward propagation. In addition, the teacher model needs to be consistent with the weighted trained model in terms of type (Seq2Seq or Seq2Edit) and tokenizer. 

\section*{Acknowledgements}
We would like to thank all the reviewers for their valuable advice to make this paper better. This research is supported by  National Science Fund for Excellent Young Scholars under Grant  62222212 and the General Program of National Natural Science Foundation of China under Grant 62376033. 

\bibliography{anthology,custom}

\begin{thebibliography}{26}
\expandafter\ifx\csname natexlab\endcsname\relax\def\natexlab#1{#1}\fi

\bibitem[{Awasthi et~al.(2019)Awasthi, Sarawagi, Goyal, Ghosh, and Piratla}]{DBLP:conf/emnlp/AwasthiSGGP19}
Abhijeet Awasthi, Sunita Sarawagi, Rasna Goyal, Sabyasachi Ghosh, and Vihari Piratla. 2019.
\newblock \href {https://doi.org/10.18653/v1/D19-1435} {Parallel iterative edit models for local sequence transduction}.
\newblock In \emph{Proceedings of the 2019 Conference on Empirical Methods in Natural Language Processing and the 9th International Joint Conference on Natural Language Processing, {EMNLP-IJCNLP} 2019, Hong Kong, China, November 3-7, 2019}, pages 4259--4269. Association for Computational Linguistics.

\bibitem[{Bryant et~al.(2019)Bryant, Felice, Andersen, and Briscoe}]{DBLP:conf/bea/BryantFAB19}
Christopher Bryant, Mariano Felice, {\O}istein~E. Andersen, and Ted Briscoe. 2019.
\newblock \href {https://doi.org/10.18653/v1/w19-4406} {The {BEA-2019} shared task on grammatical error correction}.
\newblock In \emph{Proceedings of the Fourteenth Workshop on Innovative Use of {NLP} for Building Educational Applications, BEA@ACL 2019, Florence, Italy, August 2, 2019}, pages 52--75. Association for Computational Linguistics.

\bibitem[{Bryant et~al.(2017)Bryant, Felice, and Briscoe}]{DBLP:conf/acl/BryantFB17}
Christopher Bryant, Mariano Felice, and Ted Briscoe. 2017.
\newblock \href {https://doi.org/10.18653/v1/P17-1074} {Automatic annotation and evaluation of error types for grammatical error correction}.
\newblock In \emph{Proceedings of the 55th Annual Meeting of the Association for Computational Linguistics, {ACL} 2017, Vancouver, Canada, July 30 - August 4, Volume 1: Long Papers}, pages 793--805. Association for Computational Linguistics.

\bibitem[{Bryant et~al.(2022)Bryant, Yuan, Qorib, Cao, Ng, and Briscoe}]{DBLP:journals/corr/abs-2211-05166}
Christopher Bryant, Zheng Yuan, Muhammad~Reza Qorib, Hannan Cao, Hwee~Tou Ng, and Ted Briscoe. 2022.
\newblock \href {https://doi.org/10.48550/arXiv.2211.05166} {Grammatical error correction: {A} survey of the state of the art}.
\newblock \emph{CoRR}, abs/2211.05166.

\bibitem[{Dahlmeier et~al.(2013)Dahlmeier, Ng, and Wu}]{DBLP:conf/bea/DahlmeierNW13}
Daniel Dahlmeier, Hwee~Tou Ng, and Siew~Mei Wu. 2013.
\newblock \href {https://aclanthology.org/W13-1703/} {Building a large annotated corpus of learner english: The {NUS} corpus of learner english}.
\newblock In \emph{Proceedings of the Eighth Workshop on Innovative Use of {NLP} for Building Educational Applications, BEA@NAACL-HLT 2013, June 13, 2013, Atlanta, Georgia, {USA}}, pages 22--31. The Association for Computer Linguistics.

\bibitem[{Devlin et~al.(2019)Devlin, Chang, Lee, and Toutanova}]{DBLP:conf/naacl/DevlinCLT19}
Jacob Devlin, Ming{-}Wei Chang, Kenton Lee, and Kristina Toutanova. 2019.
\newblock \href {https://doi.org/10.18653/v1/n19-1423} {{BERT:} pre-training of deep bidirectional transformers for language understanding}.
\newblock In \emph{Proceedings of the 2019 Conference of the North American Chapter of the Association for Computational Linguistics: Human Language Technologies, {NAACL-HLT} 2019, Minneapolis, MN, USA, June 2-7, 2019, Volume 1 (Long and Short Papers)}, pages 4171--4186. Association for Computational Linguistics.

\bibitem[{Ge et~al.(2018)Ge, Wei, and Zhou}]{DBLP:journals/corr/abs-1807-01270}
Tao Ge, Furu Wei, and Ming Zhou. 2018.
\newblock \href {http://arxiv.org/abs/1807.01270} {Reaching human-level performance in automatic grammatical error correction: An empirical study}.
\newblock \emph{CoRR}, abs/1807.01270.

\bibitem[{Kaneko et~al.(2020)Kaneko, Mita, Kiyono, Suzuki, and Inui}]{DBLP:conf/acl/KanekoMKSI20}
Masahiro Kaneko, Masato Mita, Shun Kiyono, Jun Suzuki, and Kentaro Inui. 2020.
\newblock \href {https://doi.org/10.18653/v1/2020.acl-main.391} {Encoder-decoder models can benefit from pre-trained masked language models in grammatical error correction}.
\newblock In \emph{Proceedings of the 58th Annual Meeting of the Association for Computational Linguistics, {ACL} 2020, Online, July 5-10, 2020}, pages 4248--4254. Association for Computational Linguistics.

\bibitem[{Lai et~al.(2022)Lai, Zhou, Zeng, Li, Li, Cao, and Su}]{DBLP:conf/acl/LaiZZLLCS22}
Shaopeng Lai, Qingyu Zhou, Jiali Zeng, Zhongli Li, Chao Li, Yunbo Cao, and Jinsong Su. 2022.
\newblock \href {https://doi.org/10.18653/v1/2022.findings-acl.254} {Type-driven multi-turn corrections for grammatical error correction}.
\newblock In \emph{Findings of the Association for Computational Linguistics: {ACL} 2022, Dublin, Ireland, May 22-27, 2022}, pages 3225--3236. Association for Computational Linguistics.

\bibitem[{Lewis et~al.(2020)Lewis, Liu, Goyal, Ghazvininejad, Mohamed, Levy, Stoyanov, and Zettlemoyer}]{DBLP:conf/acl/LewisLGGMLSZ20}
Mike Lewis, Yinhan Liu, Naman Goyal, Marjan Ghazvininejad, Abdelrahman Mohamed, Omer Levy, Veselin Stoyanov, and Luke Zettlemoyer. 2020.
\newblock \href {https://doi.org/10.18653/v1/2020.acl-main.703} {{BART:} denoising sequence-to-sequence pre-training for natural language generation, translation, and comprehension}.
\newblock In \emph{Proceedings of the 58th Annual Meeting of the Association for Computational Linguistics, {ACL} 2020, Online, July 5-10, 2020}, pages 7871--7880. Association for Computational Linguistics.

\bibitem[{Lichtarge et~al.(2020)Lichtarge, Alberti, and Kumar}]{DBLP:journals/tacl/LichtargeAK20}
Jared Lichtarge, Chris Alberti, and Shankar Kumar. 2020.
\newblock \href {https://doi.org/10.1162/tacl\_a\_00336} {Data weighted training strategies for grammatical error correction}.
\newblock \emph{Trans. Assoc. Comput. Linguistics}, 8:634--646.

\bibitem[{Liu et~al.(2019)Liu, Ott, Goyal, Du, Joshi, Chen, Levy, Lewis, Zettlemoyer, and Stoyanov}]{DBLP:journals/corr/abs-1907-11692}
Yinhan Liu, Myle Ott, Naman Goyal, Jingfei Du, Mandar Joshi, Danqi Chen, Omer Levy, Mike Lewis, Luke Zettlemoyer, and Veselin Stoyanov. 2019.
\newblock \href {http://arxiv.org/abs/1907.11692} {Roberta: {A} robustly optimized {BERT} pretraining approach}.
\newblock \emph{CoRR}, abs/1907.11692.

\bibitem[{Malmi et~al.(2019)Malmi, Krause, Rothe, Mirylenka, and Severyn}]{DBLP:conf/emnlp/MalmiKRMS19}
Eric Malmi, Sebastian Krause, Sascha Rothe, Daniil Mirylenka, and Aliaksei Severyn. 2019.
\newblock \href {https://doi.org/10.18653/v1/D19-1510} {Encode, tag, realize: High-precision text editing}.
\newblock In \emph{Proceedings of the 2019 Conference on Empirical Methods in Natural Language Processing and the 9th International Joint Conference on Natural Language Processing, {EMNLP-IJCNLP} 2019, Hong Kong, China, November 3-7, 2019}, pages 5053--5064. Association for Computational Linguistics.

\bibitem[{Ng et~al.(2014)Ng, Wu, Briscoe, Hadiwinoto, Susanto, and Bryant}]{DBLP:conf/conll/NgWBHSB14}
Hwee~Tou Ng, Siew~Mei Wu, Ted Briscoe, Christian Hadiwinoto, Raymond~Hendy Susanto, and Christopher Bryant. 2014.
\newblock \href {https://doi.org/10.3115/v1/w14-1701} {The conll-2014 shared task on grammatical error correction}.
\newblock In \emph{Proceedings of the Eighteenth Conference on Computational Natural Language Learning: Shared Task, CoNLL 2014, Baltimore, Maryland, USA, June 26-27, 2014}, pages 1--14. {ACL}.

\bibitem[{Omelianchuk et~al.(2020)Omelianchuk, Atrasevych, Chernodub, and Skurzhanskyi}]{DBLP:conf/bea/OmelianchukACS20}
Kostiantyn Omelianchuk, Vitaliy Atrasevych, Artem~N. Chernodub, and Oleksandr Skurzhanskyi. 2020.
\newblock \href {https://doi.org/10.18653/v1/2020.bea-1.16} {Gector - grammatical error correction: Tag, not rewrite}.
\newblock In \emph{Proceedings of the Fifteenth Workshop on Innovative Use of {NLP} for Building Educational Applications, BEA@ACL 2020, Online, July 10, 2020}, pages 163--170. Association for Computational Linguistics.

\bibitem[{Rothe et~al.(2021)Rothe, Mallinson, Malmi, Krause, and Severyn}]{DBLP:conf/acl/RotheMMKS20}
Sascha Rothe, Jonathan Mallinson, Eric Malmi, Sebastian Krause, and Aliaksei Severyn. 2021.
\newblock \href {https://doi.org/10.18653/v1/2021.acl-short.89} {A simple recipe for multilingual grammatical error correction}.
\newblock In \emph{Proceedings of the 59th Annual Meeting of the Association for Computational Linguistics and the 11th International Joint Conference on Natural Language Processing, {ACL/IJCNLP} 2021, (Volume 2: Short Papers), Virtual Event, August 1-6, 2021}, pages 702--707. Association for Computational Linguistics.

\bibitem[{Stahlberg and Kumar(2021)}]{DBLP:conf/bea/StahlbergK21}
Felix Stahlberg and Shankar Kumar. 2021.
\newblock \href {https://www.aclweb.org/anthology/2021.bea-1.4/} {Synthetic data generation for grammatical error correction with tagged corruption models}.
\newblock In \emph{Proceedings of the 16th Workshop on Innovative Use of {NLP} for Building Educational Applications, BEA@EACL, Online, April 20, 2021}, pages 37--47. Association for Computational Linguistics.

\bibitem[{Sun et~al.(2022)Sun, Ge, Ma, Li, Wei, and Wang}]{DBLP:conf/ijcai/0013GM0WW22}
Xin Sun, Tao Ge, Shuming Ma, Jingjing Li, Furu Wei, and Houfeng Wang. 2022.
\newblock \href {https://doi.org/10.24963/ijcai.2022/606} {A unified strategy for multilingual grammatical error correction with pre-trained cross-lingual language model}.
\newblock In \emph{Proceedings of the Thirty-First International Joint Conference on Artificial Intelligence, {IJCAI} 2022, Vienna, Austria, 23-29 July 2022}, pages 4367--4374. ijcai.org.

\bibitem[{Sun et~al.(2021)Sun, Ge, Wei, and Wang}]{DBLP:conf/acl/SunGWW20}
Xin Sun, Tao Ge, Furu Wei, and Houfeng Wang. 2021.
\newblock \href {https://doi.org/10.18653/v1/2021.acl-long.462} {Instantaneous grammatical error correction with shallow aggressive decoding}.
\newblock In \emph{Proceedings of the 59th Annual Meeting of the Association for Computational Linguistics and the 11th International Joint Conference on Natural Language Processing, {ACL/IJCNLP} 2021, (Volume 1: Long Papers), Virtual Event, August 1-6, 2021}, pages 5937--5947. Association for Computational Linguistics.

\bibitem[{Sun and Wang(2022)}]{DBLP:conf/acl/0013W22}
Xin Sun and Houfeng Wang. 2022.
\newblock \href {https://doi.org/10.18653/v1/2022.acl-short.77} {Adjusting the precision-recall trade-off with align-and-predict decoding for grammatical error correction}.
\newblock In \emph{Proceedings of the 60th Annual Meeting of the Association for Computational Linguistics (Volume 2: Short Papers), {ACL} 2022, Dublin, Ireland, May 22-27, 2022}, pages 686--693. Association for Computational Linguistics.

\bibitem[{Tarnavskyi et~al.(2022)Tarnavskyi, Chernodub, and Omelianchuk}]{DBLP:conf/acl/TarnavskyiCO22}
Maksym Tarnavskyi, Artem~N. Chernodub, and Kostiantyn Omelianchuk. 2022.
\newblock \href {https://doi.org/10.18653/v1/2022.acl-long.266} {Ensembling and knowledge distilling of large sequence taggers for grammatical error correction}.
\newblock In \emph{Proceedings of the 60th Annual Meeting of the Association for Computational Linguistics (Volume 1: Long Papers), {ACL} 2022, Dublin, Ireland, May 22-27, 2022}, pages 3842--3852. Association for Computational Linguistics.

\bibitem[{Xia et~al.(2022)Xia, Zhou, Zhang, Tang, and Li}]{xia2022chinese}
Peng Xia, Yuechi Zhou, Ziyan Zhang, Zecheng Tang, and Juntao Li. 2022.
\newblock \href {http://arxiv.org/abs/2208.00351} {Chinese grammatical error correction based on knowledge distillation}.

\bibitem[{Yannakoudakis et~al.(2011)Yannakoudakis, Briscoe, and Medlock}]{DBLP:conf/acl/YannakoudakisBM11}
Helen Yannakoudakis, Ted Briscoe, and Ben Medlock. 2011.
\newblock \href {https://aclanthology.org/P11-1019/} {A new dataset and method for automatically grading {ESOL} texts}.
\newblock In \emph{The 49th Annual Meeting of the Association for Computational Linguistics: Human Language Technologies, Proceedings of the Conference, 19-24 June, 2011, Portland, Oregon, {USA}}, pages 180--189. The Association for Computer Linguistics.

\bibitem[{Yuan and Briscoe(2016)}]{DBLP:conf/naacl/YuanB16}
Zheng Yuan and Ted Briscoe. 2016.
\newblock \href {https://doi.org/10.18653/v1/n16-1042} {Grammatical error correction using neural machine translation}.
\newblock In \emph{{NAACL} {HLT} 2016, The 2016 Conference of the North American Chapter of the Association for Computational Linguistics: Human Language Technologies, San Diego California, USA, June 12-17, 2016}, pages 380--386. The Association for Computational Linguistics.

\bibitem[{Zhang et~al.(2022{\natexlab{a}})Zhang, Li, Bao, Li, Zhang, Li, Huang, and Zhang}]{DBLP:conf/naacl/0004LBLZLHZ22}
Yue Zhang, Zhenghua Li, Zuyi Bao, Jiacheng Li, Bo~Zhang, Chen Li, Fei Huang, and Min Zhang. 2022{\natexlab{a}}.
\newblock \href {https://doi.org/10.18653/v1/2022.naacl-main.227} {Mucgec: a multi-reference multi-source evaluation dataset for chinese grammatical error correction}.
\newblock In \emph{Proceedings of the 2022 Conference of the North American Chapter of the Association for Computational Linguistics: Human Language Technologies, {NAACL} 2022, Seattle, WA, United States, July 10-15, 2022}, pages 3118--3130. Association for Computational Linguistics.

\bibitem[{Zhang et~al.(2022{\natexlab{b}})Zhang, Zhang, Li, Bao, Li, and Zhang}]{DBLP:conf/emnlp/0004ZLBLZ22}
Yue Zhang, Bo~Zhang, Zhenghua Li, Zuyi Bao, Chen Li, and Min Zhang. 2022{\natexlab{b}}.
\newblock \href {https://aclanthology.org/2022.emnlp-main.162} {Syngec: Syntax-enhanced grammatical error correction with a tailored gec-oriented parser}.
\newblock In \emph{Proceedings of the 2022 Conference on Empirical Methods in Natural Language Processing, {EMNLP} 2022, Abu Dhabi, United Arab Emirates, December 7-11, 2022}, pages 2518--2531. Association for Computational Linguistics.

\end{thebibliography}
\bibliographystyle{acl_natbib}

\appendix

\section{Training Details}
\label{sec:appendix a}

The hyper-parameters for MainGEC (BART) are listed in Table \ref{tab:parameters}.

\begin{table}[h]
\small
\centering\setlength{\tabcolsep}{4pt}
\begin{tabular*}{0.48 \textwidth}{@{\extracolsep{\fill}}lr}
\toprule
Configurations			 & Values	\\
\midrule
\multicolumn{2}{c}{\textbf{Fine-tune}} \\
\midrule
Model Architecture & BART-large \\
Number of epochs & 30 \\
Devices & 1 Tesla A800 with 80G \\
Max tokens per GPU & 20480\\
Update Frequency & 2\\
Learning rate & 3e-05 \\
\multirow{2}{*}{Optimizer} & Adam \\
  & ($\beta_1=0.9, \beta_2=0.98, \epsilon=1e-8$) \\
Learning rate scheduler & polynomial decay\\
Weight decay & 0.01 \\
Loss Function & cross entropy \\
Warmup & 2000\\
Dropout & 0.3 \\
\bottomrule
\end{tabular*}
\caption{\label{tab:parameters} Hyper-parameters values for MainGEC (BART).}
\end{table}

\section{Effect of Different Teachers}
\label{teacher}

\begin{table}[t]
\small
\centering\setlength{\tabcolsep}{4pt}
\begin{tabular*}{0.48 \textwidth}{@{\extracolsep{\fill}}lcc}
\toprule
Method & CONLL-14 & BEA-19 (dev) \\  
\midrule
GECToR (w/o teacher)&   63.4         & 52.9\\
MainGEC (w/ base teacher) & 64.9 & \textbf{55.6}\\			
MainGEC (w/ large teacher) & \textbf{65.1} & 54.9  \\						
\bottomrule
\end{tabular*}
\caption{\label{tab:teacher1} Performance of MainGEC based on teachers of different model scales.}
\end{table}

\begin{table}[t]
\small
\centering\setlength{\tabcolsep}{4pt}
\begin{tabular*}{0.48 \textwidth}{@{\extracolsep{\fill}}lcc}
\toprule
Method & CONLL-14 & BEA-19 (dev) \\  
\midrule
GECToR-L (original teacher)&   64.4         & 56.4\\
MainGEC (1st round) & 65.7 & \textbf{57.6}\\			
MainGEC (2nd round) & \textbf{66.1} & 57.4  \\						
\bottomrule
\end{tabular*}
\caption{\label{tab:teacher2} Performance of MainGEC with self-paced learning.}
\end{table}

In MainGEC, a teacher is used to quantify training weights of both granularities, which is the main contribution of this work. To investigate effect of different teacher on MainGEC, we conduct comparative experiments under two settings: 
(1) Teachers of different model scales: we use GECTOR (RoBERTa-base) and GECTOR (RoBERTa-large) as the teacher respectively for weighted training of GECTOR (RoBERTa-base). 
(2) MainGEC with self-paced learning: we use MainGEC as a stronger teacher for a new round of weighted training, {\i.e.} iterative weighted training with MainGEC. The teacher used in the second round of training is the same model scale as the teacher used in the first round but performs better in GEC. 

Table \ref{tab:teacher1} and Table \ref{tab:teacher2} present the experiment results respectively.
Experiment results show that no matter what teacher model you use, mixed-grained weights generated by them can bring improvement over the baseline, verifying effectiveness of MainGEC. Besides, this improvement is not sensitive to the choice of the teacher, either with different model sizes or with different performances in GEC. 

\end{document}